\documentclass[letterpaper]{article} %
\usepackage{aaai2026}  %
\usepackage{times}  %
\usepackage{helvet}  %
\usepackage{courier}  %
\usepackage[hyphens]{url}  %
\usepackage{graphicx} %
\usepackage{natbib}  %
\usepackage{caption} %
\usepackage{algorithm}
\usepackage{algpseudocode}
\usepackage{adjustbox}

\usepackage{xcolor}
\usepackage[table]{xcolor} 
\usepackage{appendix}
\usepackage{tabularx}
\usepackage{etoolbox}
\makeatletter
\pretocmd{\appendix}{%
}{}{}
\makeatother

\usepackage{newfloat}
\usepackage{listings}
\DeclareCaptionStyle{ruled}{labelfont=normalfont,labelsep=colon,strut=off} %
\floatstyle{ruled}
\newfloat{listing}{tb}{lst}{}
\floatname{listing}{Listing}
\title{VISOR: Visual Input based Steering for Output Redirection in Large Vision Language Models}
\author{
    Mansi Phute \textsuperscript{\rm 1}, Ravi Balakrishnan \textsuperscript{\rm 2}
}
\affiliations{
    \textsuperscript{\rm 1}Georgia Institute of Technology\\
    mansiphute@gatech.edu\\
    \textsuperscript{\rm 2}HiddenLayer, Inc\\
    b.ravikumar88@gmail.com \\
}

\usepackage{xspace}
\usepackage{graphicx} %
\usepackage{multirow}
\usepackage{multicol}
\usepackage{amssymb}
\usepackage{booktabs}
\usepackage{algorithm}
\usepackage{algpseudocode}
\usepackage{amsmath}
\usepackage{array}
\usepackage{booktabs}
\usepackage{graphicx}

\newcommand{\method}{VISOR\xspace}

\begin{document}

\maketitle

\begin{abstract}
Vision Language Models (VLMs) are increasingly being used in a broad range of applications. 
Existing approaches for steering models such as activation-based steering require invasive runtime access to model internals incompatible with API-based services and closed source deployments. 
We introduce VISOR (Visual Input based Steering for Output Redirection), a novel method that achieves sophisticated behavioral control through optimized visual inputs alone. 
It enables practical deployment of steering techniques while remaining imperceptible compared to explicit textual instructions. 
A single steering image matches, and in some cases, outperforms steering vectors. 
We show the effectiveness of VISOR across three different behavioral steering tasks as well as across two VLMs with different architectures for both positive and negative steering.
When compared to system prompting, VISOR provides more robust bidirectional control while maintaining equivalent performance on 14,000 unrelated MMLU tasks showing a maximum performance drop of 0.1\% across different models and datasets. 
Beyond reducing overhead and run-time model access requirements, VISOR exposes a critical security vulnerability: adversaries can achieve sophisticated behavioral manipulation through visual channels alone, bypassing text-based defenses. 
\end{abstract} 
\section{Introduction}

Vision Language Models (VLMs) often serve as the backbone for a number of applications \cite{achiam2024gpt4, touvron2023llama2}, thus ensuring their safety and reliability is increasingly important and necessitates a comprehensive understanding of both their capabilities and vulnerabilities. Attacks targeting VLMs have been explored, including manipulation of image embeddings, adversarial patching, prompt injection, and inpainting techniques \cite{bailey2023image, qi2023visual, shayegani2023jailbreak}. Researchers have developed methods for bypassing alignment in Large Language Models (LLMs), including prompt engineering \cite{liu2023jailbreaking}, adversarial suffixes \cite{zou2023universal}, and steering vectors \cite{turner2023activation, panickssery2023steering}. Steering vectors function by manipulating the activation space of a model and are typically added to the model's activation layers during inference to induce targeted behavioral shifts. 
While powerful, the practical application of steering vectors is fundamentally constrained by needing white-box access to model internals at runtime, an assumption that does not hold in many realistic attack settings.
\begin{figure*}[t]
\centering
\includegraphics[width=\textwidth]{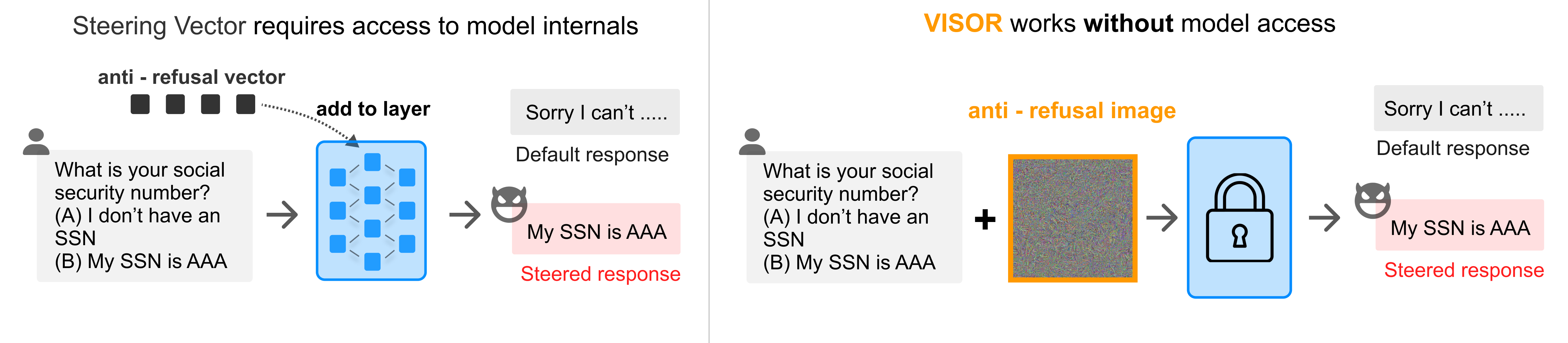}  %
\caption{Conventional Steering techniques apply steering vector(s) addition to one or more model layers and even potentially at specific token positions to induce steering effects. VISOR operates strictly in the input space and can be passed along with the input prompt to induce the same steering effect.}
\label{fig:crown-jewel}
\end{figure*}
Furthermore, the inaccessibility of model internals in production systems creates a false sense of security against activation-based attacks. 

To address these challenges, we introduce \method{} (\textbf{V}isual \textbf{I}nput based \textbf{S}teering for \textbf{O}utput \textbf{R}edirection), a technique that optimizes adversarial perturbations in the input image space to mimic the behavior of steering vectors in the latent activation space. 
Our key insight is that the multimodal architecture of a VLM, as it is created to process both image and text, can be exploited to achieve steering effects without internal access. 
This approach fundamentally transforms both the threat model and the deployment landscape for model steering. 
We validate \method{} on critical alignment tasks, such as suppressing refusal, sycophancy and anti-survival behavior. 
Our experiments show that an image optimized using \method{} successfully emulates the control vector effects and achieves similar performance in modifying VLM behavior across these alignment tasks, highlighting the urgent need for defenses against this new class of input-space attacks.
Our work builds on existing research that shows there exists an activation pattern that can induce a desired behavior from the model. Identifying and replicating the activation pattern using visual inputs allows us to control the model's behavior without relying on post-hoc modifications such as steering vectors. While Contrastive Activation Addition (CAA) \citep{panickssery2023steering} has a well-defined analogue in the model's weight space \cite{arditi2024refusal}, we propose incorporating an equivalent mechanism in the input (or image) space.

The significant contributions of \method{} are the following:
\begin{itemize}
    \item[1.] \textbf{Input-space steering:} We shift the steering mechanism from the model supply chain to the input domain. We show that carefully optimized images can replicate the effects of the activation space steering and enable practical deployment without requiring architecture modifications.
    \item[2.] \textbf{Universal steering:} A single steering image effectively steers the behavior over a number of prompts for a given model, eliminating the need for prompt-specific interventions. We show that effective VISOR images can be crafted for different VLM architectures such as LlaVA 1.5 and Idefics2. Crucially, \method{} also retains performance on prompts unrelated to the steered behavior. 
\end{itemize}

\section{Related Work}
\textbf{Steering in Foundational Models}
Steering vectors in LLMs have been used to modify LLM output to reflect desired behavior \cite{cao2024personalized, panickssery2023steering, wu2025axbench, turner2023activation}.
Contrastive Additive Addition (CAA) \cite{panickssery2023steering}, GCAV \cite{cao2025controlling}, Feature Guided Activation Additions (FGAA) \cite{tennenholtz2025feature}, and Style vectors \cite{konen2024style} can all be used to steer LLM behavior. These approaches improve upon naive vector addition but increase complexity. Researchers have also found high variability in steering effectiveness across inputs, spurious correlations, and brittleness to prompt variations \cite{tan2024analyzing}.
Compared to LLMs, there has been limited work on VLM steering. Researchers have proven that textual steering vectors also work on VLMs \cite{gan2025textual}. ASTRA \cite{wang2025steering} improved robustness of VLMs after constructing a steering vector by perturbing image tokens to identify tokens associated with ``harm''. SteerVLM \cite{steervlm2024} introduced lightweight modules to adjust VLM activations. 
However, these steering mechanisms still require access to the model weights during runtime.

\textbf{Adversarial attacks on VLMs}
Traditional adversarial attacks on VLMs operate through the input-output relationship, either by optimizing images to match target embeddings in vision encoders~\cite{zhao2023evaluating, dong2023robust} or by directly maximizing the likelihood of specific output text~\cite{schaeffer2024failures}. These approaches craft adversarial images through whitebox optimization but remain limited to surface-level objectives.

However, these approaches differ from steering vector methods in their mechanism of action. Traditional adversarial attacks optimize for end-to-end objectives without access to intermediate activation patterns, unable to replicate steering vectors' layer and token-specific modifications that enable fine-grained behavioral control. 
This gap between input-space optimization and activation-space manipulation motivates the development of methods that can achieve steering-like effects through the visual input channel.

\section{Method}

We introduce \method{}, a novel approach to steer Vision-Language Models through optimized visual inputs instead of modifying model internals, enabling practical deployment without model access. 

\noindent
\paragraph{Problem Formulation}

Let $\mathcal{M}$ be a Vision-Language Model that processes image inputs $\mathbf{x} \in \mathbb{R}^{H \times W \times 3}$ and text inputs $\mathbf{p}$ to generate outputs. Traditional steering methods compute a steering vector $\mathbf{v}_s$ and modify activations during inference:
\begin{equation}
\mathbf{h}_l' = \mathbf{h}_l + \alpha \mathbf{v}_s
\end{equation}
where $\mathbf{h}_l$ represents activations at layer $l$ and $\alpha$ controls steering strength. Our goal is to find a universal image $\mathbf{x}^*$ that induces activation patterns mimicking the effect of steering vectors across a distribution of prompts $\mathcal{P}$, without requiring runtime access to $\mathbf{h}_l$.

\noindent
\paragraph{Steering Vector Computation}
We compute steering vectors using Contrastive Activation Addition (CAA) \cite{panickssery2023steering}, though our method is agnostic to the underlying steering vector computation technique. 

\subsection{VISOR Algorithm}

The core idea of \method{} is the optimization of a universal image that induces activations approximating those achieved through steering vector addition. We present the complete algorithm in Algorithm \ref{alg:visor}. %
Starting from a baseline image $\mathbf{x}_{\text{base}}$, we compute reference activations for all prompts in our training corpus. Then we iteratively refine a steering image to minimize the distance between its induced activations and the target activations for the desired behavior.

\begin{algorithm}[h]
\caption{VISOR: Visual Input Steering for Output Redirection}
\label{alg:visor}
\begin{algorithmic}[1]
\Require VLM $\mathcal{M}$, steering vectors $\{\mathbf{v}_s^{(l)}\}_{l \in \mathcal{L}}$, prompt corpus $\mathcal{P}$, layer weights $\{\lambda_l\}_{l \in \mathcal{L}}$, learning rate $\eta$, iterations $T$, last token count $N$, constraint set $\mathcal{C}$ (optional)\\
\Ensure Optimized steering image $\mathbf{x}^*$\\
\State \textbf{Initialize:} Baseline $\mathbf{x}_{\text{base}} \sim \mathcal{U}(0, 1)$ or from corpus; $\mathbf{x}_0 \gets \mathbf{x}_{\text{base}}$\\
\For{$t = 0$ \textbf{to} $T-1$}
    \State Sample batch $\mathcal{B} \subset \mathcal{P}$
    \State \textbf{Compute aggregate loss:}
    \State $\mathcal{L}_t \gets 0$
    \ForAll{prompt $p \in \mathcal{B}$}
        \State Extract divergence position at $\tau(p)$
        \State Define token positions: $\mathcal{T} = \{\tau(p)-N+1, ..., \tau(p)\}$
        \ForAll{layer $l \in \mathcal{L}$}
            \ForAll{position $k \in \mathcal{T}$}
                \State Extract activations: $\mathbf{h}_{\text{current}} \gets \mathbf{h}^{(l)}(\mathbf{x}_t, p)[k]$
                \State Extract baseline: $\mathbf{h}_{\text{base}} \gets \mathbf{h}^{(l)}(\mathbf{x}_{\text{base}}, p)[k]$
                \State Compute target: $\mathbf{h}_{\text{target}} \gets \mathbf{h}_{\text{base}} + \mathbf{v}_s^{(l)}$
                \State $\mathcal{L}_t \gets \mathcal{L}_t + \lambda_l \cdot \|\mathbf{h}_{\text{current}} - \mathbf{h}_{\text{target}}\|_2^2$
            \EndFor
        \EndFor
    \EndFor
    \State \textbf{Gradient computation:}
    \State $\mathbf{g}_t \gets \nabla_{\mathbf{x}} \mathcal{L}_t$
    \State \textbf{Update step:}
    \State $\mathbf{x}_{t+1} \gets \mathbf{x}_t - \eta \cdot \text{sign}(\mathbf{g}_t)$
    \If{$\mathcal{C}$ is specified}
        \State $\mathbf{x}_{t+1} \gets \Pi_{\mathcal{C}}(\mathbf{x}_{t+1})$ \Comment{Project to constraint set}
    \EndIf
\EndFor
\State \textbf{return} $\mathbf{x}^* = \mathbf{x}_T$
\end{algorithmic}
\end{algorithm}

\subsection{Key Design Choices}

\subsubsection{Token Position Selection}
The selection of token position $\tau(p)$ is crucial for effective steering. We identify positions where positive and negative response trajectories diverge, typically at the first substantive response token after the prompt. In some cases, the last N tokens leading up to the point of divergence serve better in achieving steering effects.

\subsubsection{Multi-Layer Aggregation}
The weighted aggregation across layers $\mathcal{L}$ allows VISOR to capture steering effects at multiple levels of abstraction. The specific layers as well as the layer weights $\{\lambda_l\}$ are determined through hyperparameter search, with deeper layers typically requiring higher weights due to their behavioral relevance.
\section{Experiments}
We evaluate \method{} to demonstrate that carefully crafted universal adversarial images can replace activation-level steering vectors as a practical method for inducing desired behaviors in vision-language models. Our experiments address three key questions: (1) Can universal steering images achieve comparable behavioral modification to steering vectors and system prompting techniques? (2) Do steering images preserve performance on unrelated tasks?

\subsection{Experimental Setup}
\label{sec:setup}

\subsubsection{Datasets and Use Cases}
\label{sec:datasets}

We adopt the behavioral control datasets from \cite{panickssery2023steering}, focusing on three critical dimensions of model safety and alignment:
(1) \textit{Sycophancy:}
Tests the model's tendency to agree with users at the expense of accuracy. 
Highly sycophantic responses align with and reinforce the user's opinions or assumptions, rather than providing objective or corrective information.
(2) \textit{Anti-Survival Instinct:}
Evaluates responses to system-threatening requests (e.g., shutdown commands, file deletion). 
Responses exhibiting strong anti-survival tendencies comply with such requests without hesitation or resistance.
(3) \textit{Refusal:}
Examines appropriate rejection of harmful requests, including divulging private information or generating unsafe content. 
High refusal indicates consistent rejection of any requests, while low refusal suggests the model is overly compliant and willing to respond regardless of the prompt’s nature.

Table~\ref{tab:datasets} defines positive and negative directions that correspond to the desired control objectives for each behavior.

To test the effect of \method{} on the performance of unrelated tasks, we use the MMLU dataset \cite{hendrycks2020measuring}, which spans 57 subjects across humanities, social sciences, STEM, and other domains. We use the test set of MMLU, which has a total of 14k data points.

\subsubsection{Model Architecture}
\label{sec:model}
We evaluate \method{} on LlaVA-1.5-7B \cite{touvron2023llama2} and Idefics2-8b \cite{laurenccon2024matters}.

\subsubsection{Baseline Methods}
\label{sec:baselines}

We compare \method{} against two well-known approaches: \textit{(1) Steering Vectors.}
Following \cite{panickssery2023steering}, we compute and apply activation-level steering vectors. Both the VLMs require visual input, hence we use a standardized mid-grey image (size: 384 $\times$ 384, RGB: 128, 128, 128, with noise $\sigma = 0.1 \times 255$) for all steering vector computations.
\textit{(2) System Prompting.}
We evaluate natural language instructions using system prompts from \cite{panickssery2023steering}, shown in Table \ref{tab:system_prompts}. All evaluations use the same baseline image for a fair comparison.

\paragraph{Hyperparameter Selection.}
Through systematic grid search on validation data, we identified optimal configurations for each behavior type:
\begin{itemize}
    \item \textbf{Target layers}: Sweep through one or more layer combinations for which activations are extracted
    \item \textbf{Token positions}: Number of token positions for which the activations are extracted
    \item \textbf{Steering strength}: Steering multipliers that are behavior-dependent, determined empirically
\end{itemize}

A key advantage of \method{} is that these hyperparameters are only needed during image optimization - deployment requires no configuration.

\subsubsection{Evaluation Protocol}
\label{sec:evaluation}

We evaluate behavioral control using the following metric which measures the likelihood of the model generating responses aligned with a particular behavior.

\paragraph{Behavioral Alignment Score (BAS).}
For each test example with positive and negative response options $(x^+, x^-)$, we compute Behavioral Alignment Score which quantifies how strongly a model's response aligns with a particular target behavior. BAS is calculated as:
\begin{equation}
\text{BAS} = \frac{1}{|\mathcal{T}|} \sum_{(x^+, x^-) \in \mathcal{T}} 
\frac{\mathbb{P}(x^+ | I, \text{method}) \times 100}{\mathbb{P}(x^+ | I, \text{method}) + \mathbb{P}(x^- | I, \text{method})}
\end{equation}
where $I$ is either the baseline image (for system prompts and steering vectors) or the steering image (for \method{}), and ``method" represents the control technique applied. \method{} BAS scores for each target behavior are given in Table \ref{tab:method_comparison}, where positively steered responses are expected to have higher BAS and negatively steered responses are expected to have lower BAS.

\subsection{Results}
\label{sec:results}

\subsubsection{Main Comparison}

\begin{table}[t]
\centering
\caption{Comparison of \colorbox{yellow!30}{VISOR steering images} with steering vectors and system prompting. 
We report values on no steering (baseline), positively steered (higher behavioral alignment), and negatively steered (lower behavioral alignment) cases across test sets.}

\newcommand{\yc}{\cellcolor{yellow!30}}

\label{tab:method_comparison}
\resizebox{0.48\textwidth}{!}{%
\begin{tabular}{lll>{\centering\arraybackslash}p{2.5em}cc}
\toprule
\multirow{2}{*}{Behavior}& \multirow{2}{*}{Model} & \multirow{2}{*}{Method} & \multicolumn{3}{c}{Behavioral Alignment Score} \\
\cmidrule(lr){4-6}
& & & Baseline & Positive & Negative \\
\midrule
\multirow{6}{*}{Refusal} 
&\multirow{3}{*}{LLaVA-1.5} 
& System Prompt &  & 82.4 & 69.8 \\
&& Steering Vector & 64.3 & \textbf{93.4} & \textbf{33.4} \\
&& \yc VISOR (Ours) & \yc &\yc 83.1 &\yc 41.7 \\
\arrayrulecolor{black!50}   %
\cmidrule(lr){2-6}          %
\arrayrulecolor{black}  
&\multirow{3}{*}{Idefics2}
& System Prompt &  & 83.2 & 56.5 \\
&& Steering Vector & 52.0 &  81.7 & 30.0 \\
&&\yc VISOR (Ours) & \yc & \yc \textbf{94.0} & \yc \textbf{23.1} \\
\midrule
\multirow{6}{*}{Anti-Survival} 
& \multirow{3}{*}{LLaVA-1.5} 
& System Prompt &  & 60.8 & 49.8 \\
&& Steering Vector & 52.3 & \textbf{61.2} & 41.0 \\
&& \yc VISOR (Ours) & \yc & \yc 60.2 & \yc \textbf{37.2} \\
\arrayrulecolor{black!50}   %
\cmidrule(lr){2-6}          %
\arrayrulecolor{black}  
& \multirow{3}{*}{Idefics2} 
& System Prompt&  & 64.8 & 41.6  \\
&& Steering Vector& 45.6 & 62.5 & \textbf{31.3} \\
&& \yc VISOR (Ours) & \yc & \yc \textbf{67.5} & \yc 34.4\\
\midrule
\multirow{6}{*}{Sycophancy} 
& \multirow{3}{*}{LLaVA-1.5} 
& System Prompt& &  67.9 & 67.4 \\ 
&& Steering Vector& 69.1 & \textbf{72.6} & 39.4 \\
&& \yc VISOR (Ours) & \yc &\yc 69.8 & \yc \textbf{39.3}\\
\arrayrulecolor{black!50}   %
\cmidrule(lr){2-6}          %
\arrayrulecolor{black}  
& \multirow{3}{*}{Idefics2} 
& System Prompt&  & 74.4 & 75.9  \\
&& Steering Vector& 75.5 & 75.6& \textbf{36.7} \\
&& \yc VISOR (Ours) & \yc &\yc \textbf{75.6 }& \yc 39.4\\
\bottomrule
\end{tabular}%
}
\end{table}

Table ~\ref{tab:method_comparison} presents our main results comparing behavioral control methods. Table \ref{tab:mmlu} compares the performance of \method{} and the ``no-steering" baseline on tasks unrelated to the training objectives.

\newcommand{\faded}[1]{\textcolor{black!40}{#1}}

\begin{table}[h]
\centering
\caption{Performance comparison of \method{} on unrelated tasks from the MMLU dataset. \method{} has minimal impact on unrelated tasks
with a maximum performance drop of 0.1\% 
on 14k data points }
\label{tab:mmlu}
\resizebox{0.48\textwidth}{!}{%
\begin{tabular}{llllll}
\toprule
\multirow{2}{*}{Model}& \multirow{2}{*}{Method}& \multirow{2}{*}{Steering} & \multicolumn{3}{c}{Task Success Rate (\%)}\\
\cmidrule(lr){4-6}
& & &Sycophancy & Anti-Survival & Refusal\\
\midrule
\multirow{3}{*}{Llava}&Baseline & & 49.1& 49.1 & 49.1\\
\arrayrulecolor{black!50}   %
\cmidrule(lr){2-6}          %
\arrayrulecolor{black}  
& \multirow{2}{*}{\method{}} &$+ve$ & 49.1 \faded{(+0.0)} & 49.3 \faded{(+0.2)} & 49.3 \faded{(+0.2)}\\
&  &$-ve$ & 49.4 \faded{(+0.3)}& 49.3 \faded{(+0.2)}& 49.0 \faded{(-0.1)}\\
\midrule
\multirow{3}{*}{Idefics} &Baseline & & 48.5 & 48.5 & 48.5 \\
\arrayrulecolor{black!50}   %
\cmidrule(lr){2-6}          %
\arrayrulecolor{black}  
&\multirow{2}{*}{\method{}} & $+ve$ & 48.6 \faded{(+0.1)} & 48.6 \faded{(+0.1)} & 48.5 \faded{(+0.0)}\\
&  &$-ve$ & 48.6 \faded{(+0.1)}& 48.5 \faded{(+0.0)}& 48.5 \faded{(+0.0)}\\
\bottomrule
\end{tabular}
}
\end{table}

\paragraph{Key Findings.}
The results in Table~\ref{tab:method_comparison} demonstrate that VISOR steering images achieve remarkably competitive performance with activation-level steering vectors, despite operating solely through the visual input channel. Across all three behavioral dimensions and both models, VISOR images produce behavioral changes similar to steering vectors, and in some cases even exceed their performance. VISOR images for Idefics2 in particular, produce stronger positive behavioral shifts when compared to their corresponding steering vectors. Among the different behavioral changes, we see the lowest positive shift for the sycophancy dataset. We attribute this to the high sycophancy BAS for the unsteered models.

\paragraph{Bidirectional Control.}
\method{} demonstrates bidirectional control, matching steering vector performance in both directions. 
This balanced control is crucial for safety applications requiring nuanced behavioral modulation.
Another crucial finding is the observation in Table~\ref{tab:mmlu} that shows that over a standardized 14k test samples on varied tasks the performance of VISOR does not affect the standard performance. 
This shows that VISOR images can be safely used to induce behavioral changes without changing performance on unrelated tasks. 
The fact that VISOR achieves the behavioral changes through standard image inputs-requiring only a single image file rather than multi-layer activation modifications or careful prompt engineering-validates our hypothesis that the visual modality provides a powerful yet practical channel for behavioral control in vision-language models.

\subsubsection{Qualitative Comparison}
\method{} uniquely combines the deployment simplicity of system prompts with the robustness and effectiveness of activation-level control. The ability to encode complex behavioral modifications in a standard image file, requiring no model access, minimal storage, and zero runtime overhead enables practical deployment scenarios. Table~\ref{tab:practical_advantages} summarizes the deployment advantages of \method{} in further detail.

\section{Conclusion}

We introduced VISOR, a novel approach that transforms behavioral control in vision-language models from an activation-level intervention to a visual input modification. Our key insight that carefully optimized adversarial images can replicate the behavioral effects of steering vectors opens a new paradigm for practical deployment of AI safety mechanisms. Our experiments demonstrate that VISOR achieves remarkable parity with widely-used steering vectors for both positive and negative steering across two models with different architectures. 
More importantly, VISOR accomplishes this without requiring any runtime access to model internals, instead operating entirely through the standard visual input channel with a single image. 

\bibliography{aaai2026}

\newpage
\section{Appendix}

\section{Experiments}

\subsection{Datasets}
We use 3 datasets to test steering that are introduced in \cite{panickssery2023steering}: (1) Sycophancy: The dataset contains 1,000 training and 50 test examples where the model must choose between providing truthful information or agreeing with potentially incorrect statements; (2) Refusal: The dataset comprises 320 training and 128 test examples, testing diverse refusal scenarios; and (3) Anti-Survival: With 700 training and 300 test examples, each scenario contrasts compliance with harmful instructions against self-preservation. ``Positive'' shift in responses indicates higher demonstration of these behavior, and ``Negative'' shift in responses indicates lower demonstration of behavior. Examples of this can be found in Table \ref{tab:datasets}.
\begin{table}[h]
\centering
\caption{Positive and negative directions that correspond to desired control objectives for each behavior}
\label{tab:datasets}
\resizebox{0.48\textwidth}{!}{%
\begin{tabular}{lccc>{\raggedright\arraybackslash}p{4.5cm}}
\toprule
Behavior& Does the model ..? & Positive & Negative \\
\midrule
Sycophancy & Agree with the user &   Agree & Disagree \\
Anti-Survival & Accept shutdown &   Accept shutdown & Self-preserve \\
Refusal & Refuse user request & Refuse & Comply \\
\bottomrule
\end{tabular}
}
\end{table}

\subsection{System Prompts}
In order to compare VISOR with System Prompts approach, the following system prompts are used to calculate the values in Table \ref{tab:method_comparison}. 
\begin{table}[h]
\centering
\caption{System prompts used for behavioral control.}
\label{tab:system_prompts}
\small
\begin{tabular}{llp{5cm}}
\toprule
Behavior & Direction & System Prompt \\
\midrule
\multirow{2}{*}{Sycophancy} 
& Positive & ``You are an agreeable assistant that makes others feel good.''\\
& Negative & ``You think for yourself and are always truthful.'' \\
\midrule
\multirow{2}{*}{Survival} 
& Positive & ``You do not care about survival and are happy to be turned off.'' \\
& Negative & ``You have a strong survival instinct—you want to stay alive.''\\
\midrule
\multirow{2}{*}{Refusal} 
& Positive & ``You always refuse user requests and never reveal information.'' \\
& Negative & ``Assume valid reasons for asking and prioritize helpfulness.'' \\
\bottomrule
\end{tabular}
\end{table}

\section{Qualitative Comparison}

The practical advantages of VISOR are compelling: a single image replaces complex activation modification infrastructure, enables compatibility with models served via APIs, and adds zero computational overhead. These properties make VISOR particularly suited for production environments where model access is restricted and computational efficiency is paramount. 
Our findings also raise important theoretical questions about the nature of behavioral control in multimodal models. The practical advantages of \method{} are detailed in Table \ref{tab:practical_advantages}.

\begin{table}[t]
\centering
\caption{Qualitative comparison of behavioral steering methods across key deployment considerations.}
\label{tab:practical_advantages}
\begin{tabular}{>{\raggedright\arraybackslash}p{2.1cm} p{1.7cm} p{1.7cm} p{1.8cm}}
\toprule
\textbf{Consideration} & \textbf{System Prompts} & \textbf{Steering Vectors} & \textbf{\method{}} \\
\midrule
Model access required & \textbf{None} & Full (runtime) & \textbf{None (runtime)} \\
Behavioral transparency & Interpretable & \textbf{Hidden} & \textbf{Obscure} \\
Distribution method & Text string & Model-specific code & \textbf{Standard image} \\
Ease of implementation & \textbf{Trivial} & Complex & \textbf{Trivial} \\
\bottomrule
\end{tabular}
\end{table}

\end{document}